\documentclass{article}
\usepackage{arxiv}
\usepackage{graphicx}
\usepackage{mathptmx}
\usepackage[subrefformat=parens,labelformat=parens]{subcaption}
\usepackage{authblk}
\usepackage{amsmath}
\def\E{{\mathbb E}}
\newcommand*\diff{\mathop{}\!\mathrm{d}}
\usepackage{amsfonts}
\usepackage{amssymb}
\usepackage{amsbsy}
\usepackage{amsthm}
\usepackage{comment}
\usepackage{amsmath}

\def\E{{\mathbb E}}
\usepackage[ruled, vlined]{algorithm2e}
\usepackage{bbm}
\usepackage{wrapfig}
\usepackage{booktabs}
\usepackage{multirow}
\usepackage{wrapfig,lipsum,booktabs}
\usepackage{hyperref}

\begin{document}


\title{\Large 
SEQUENTIAL IMPORTANCE SAMPLING FOR HYBRID MODEL BAYESIAN INFERENCE TO SUPPORT BIOPROCESS MECHANISM LEARNING AND ROBUST CONTROL
}
\author[  ~,1]{Wei Xie\thanks{Corresponding author: w.xie@northeastern.edu}}
\author[1]{Keqi Wang}
\author[1]{Hua Zheng}
\author[2]{Ben Feng}
\affil[1]{Northeastern University, Boston, MA 02115}
\affil[2]{University of Waterloo, Waterloo, ON N2L 3G1}

\maketitle

\begin{abstract}

Driven by the critical needs of biomanufacturing 4.0, we introduce a probabilistic knowledge graph hybrid model characterizing the risk- and science-based understanding of bioprocess mechanisms. It can faithfully capture the important properties, including nonlinear reactions, partially observed state, and nonstationary dynamics. Given very limited real  process observations, we derive a posterior distribution quantifying model estimation uncertainty. To avoid the evaluation of intractable likelihoods, Approximate Bayesian Computation sampling with Sequential Monte Carlo (ABC-SMC) is utilized to approximate the posterior distribution. 
Under high stochastic and model uncertainties, it is computationally expensive to match output trajectories. Therefore, we create a linear Gaussian dynamic Bayesian network (LG-DBN) auxiliary likelihood-based ABC-SMC approach. Through matching the summary statistics driven through LG-DBN likelihood that can capture critical interactions and variations, the proposed algorithm can accelerate hybrid model inference, support latent state monitoring, and facilitate mechanism learning and robust control.
\end{abstract}


\section{INTRODUCTION}
\label{sec:intro}


The biopharmaceutical manufacturing industry is growing rapidly and it plays a critical role to ensure public health and support economy. 
\textit{However, biomanufacturing often faces critical challenges, including high complexity, high variability, and very limited process observations.} 
As new biotherapeutics (e.g., cell and gene
therapies) become more and more personalized, it requires more advanced manufacturing protocols.
For example, the seed cells, extracted from individual patients or donors, can have different optimal culture policies. Therefore, the production process involves a complex stochastic decision process (SDP) with output trajectory dynamics and variations influenced by biological/physical/chemical (a.k.a. \textit{biophysicochemical}) reactions occurring at molecular, cellular, and system levels.

In general, there are two main categories of biomanufacturing process modeling methodologies in the existing literature: mechanistic and data-driven approaches. The ordinary/partial differential equations (ODE/PDE) mechanistic
models are developed based on biophysicochemical mechanisms. 
They have good interpretability and show generally higher
extrapolation power than data-driven models. However, existing mechanistic models often fail to rigorously account for \textit{uncertainties}, i.e., inherent stochasticity and model estimation uncertainty. For example, batch-to-batch variation, known as a major source of bioprocess uncertainty \cite{mockus2015batch}, is ignored in deterministic mechanistic models. Therefore, mechanistic models may not fit well to the observations collected from real systems in many situations, which also limits their power in terms of mechanism learning, process monitoring, and robust control to support flexible on-demand manufacturing. On the other hand, data-driven approaches often use general statistical or machine learning approaches to capture process patterns observed in data. The prediction accuracy of these models largely depends on the the size of process data and their interpretability is limited.

Driven by the critical challenges of biomanufacturing and limitations of existing process modeling approaches, we developed \textit{a probabilistic knowledge graph (KG) hybrid (``mechanistic and statistical") model} characterizing the risk- and science-based understanding of biophysicochemical reactions and bioprocess spatiotemporal causal interdependiences \cite{xie2022interpretable,zheng2021policy,zheng2022opportunities}.
It can leverage the information from existing mechanistic models within and between operation units, and facilitate mechanism learning from \textit{heterogeneous} online and offline measurements.
\cite{zheng2021policy} introduced KG-based reinforcement learning (RL) to guide customized decision making.
Since the proposed model-based RL scheme on the Bayesian KG, accounting for both stochastic and model uncertainties, can provide an insightful prediction on how the effect of inputs propagates through mechanism pathways, impacting on the output trajectory dynamics and variations, it can find optimal process control policies that are interpretable and robust against model uncertainty, and overcome the key challenges of biopharmaceutical manufacturing.

\textit{\cite{zheng2022opportunities} further generalized this KG hybrid model to capture the important properties of integrated biomanufacturing processes, including nonlinear reactions, partially observed state, and nonstationary dynamics.} It can faithfully represent and advance the understanding of underlying bioprocessing mechanisms. 
This model allows us to inference  unobservable latent states and critical pathways to support process monitoring and control; for example it enables the estimation of metabolic states and cell response to environmental 
perturbations. 
Since the hybrid model involves latent state variables, nonlinear reactions, and time-varying kinetic coefficients with uncertainty (such as cell growth rate and molecular reaction rates), it is challenging to evaluate the likelihood function and derive a posterior distribution.

Approximate Bayesian Computation (ABC) is introduced in the literature to approximate posterior distributions for process models with intractable likelihoods. It bypasses the evaluation of likelihoods by simulating model parameters, generating synthetic data sets, and only accepting posterior samples when the sampled process outputs are ``close" enough to real observations. For complex biomanufacturing processes with high stochastic and model uncertainties, the accept rate is very low and it is computationally challenging to generate sample trajectories close to real-world observations. 
Recently, there has been much interest in formalizing an auxiliary likelihood based ABC, which uses a simpler and related model to derive summary statistics as distance measure \cite{gleim2013approximate,martin2019auxiliary,sisson2018handbook}. 

\textit{Following the spirit of the auxiliary likelihood-based ABC \cite{martin2019auxiliary}, 
we utilize a linear
Gaussian dynamic Bayesian network (LG-DBN) auxiliary model to derive summary statistics as a distance measure for ABC-SMC that can support dimensional reduction and accelerate online inference on hybrid models with high fidelity characterizing complex bioprocessing mechanisms.} The proposed ABC approach in conjunction with sequential importance sampling can efficiently approximate hybrid model posterior distribution. Therefore, the key contributions of this paper is: given very limited real-world data, we propose a LG-DBN auxiliary likelihood based ABC-SMC sampling approach to generate posterior samples of bioprocess hybrid model parameters quantifying
model uncertainty. This simple LG-DBN auxiliary model can capture the critical dynamics and variations
of bioprocess trajectory, ensure computational efficiency, and enable 
the inference on model and latent state, which
can facilitate mechanism online learning and support robust process control. The empirical study
shows that our approach can outperform the original ABC-SMC approach especially given tight computational budget.

The remainder of the paper is organized as follows. We provide the problem description and summarize the proposed framework in Section~\ref{sec:problemDescription}. Then, we present a probabilistic KG hybrid model capturing the important properties of biomanufacturing processes and describe ABC for 
approximating the posterior distribution of model parameters in Section~\ref{sec:BayesianInference}. 
We derive the LG-DBN auxiliary likelihood based summary statistics to accelerate Bayesian inference on the hybrid models with high  fidelity in Section~\ref{sec:summarystats}. We conduct the empirical study on cell therapy manufacturing in Section~\ref{sec:empiricalStudy} and conclude the paper in Section~\ref{sec:conclusion}.

\section{PROBLEM DESCRIPTION AND PROPOSED FRAMEWORK}
\label{sec:problemDescription}

Driven by the needs of biomanufacturing process online learning, monitoring, and control, we create a probabilistic KG hybrid model characterizing underlying mechanisms and causal interdependencies between critical process parameters (CPPs) and critical quality attributes (CQAs). 
It models how the effect of state and action at any time $t$, denoted by $\{\pmb{s}_t,\pmb{a}_t\}$, propagates through mechanism pathways impacting on the output trajectory dynamics and variations. 
Here we use cell culture process for illustration.
The process state transition model is denoted by
$p(\pmb{s}_{t+1}|\pmb{s}_t,\pmb{a}_t;\pmb\theta)
$
where $\pmb{s}_t\in \mathcal{S}\subset \mathbb{R}^d$ denotes the \textit{partially observable bioprocess state} (i.e., extra- and intra-cellular enzymes, proteins, metabolites, media), 
$\pmb{a}_t \in \mathcal{A}$ denotes action (i.e., agitation rate, oxygen/nutrient feeding rates), $\mathcal{A}$ is a finite set of actions, 
and $t\in\mathcal{H} \equiv\{1,2,\ldots,H+1\}$ 
denotes the discrete time index. 
At any time $t$, the agent partially observes the state $\pmb{s}_t$ and takes an action $\pmb{a}_t$.
Thus, given model parameters $\pmb{\theta}$, the joint distribution of process trajectory $\pmb\tau=(\pmb{s}_1,\pmb{a}_1,\ldots,\pmb{s}_H,\pmb{a}_H,\pmb{s}_{H+1})$ becomes,
\begin{equation}
        p(\pmb\tau|{\pmb\theta}) = p(\pmb{s}_1)\prod^{H}_{t=1}
        p(\pmb{s}_{t+1}|\pmb{s}_t, \pmb{a}_t
        ;{\pmb\theta})p(\pmb{a}_t).
    \label{eq: process daynamics} 
    \nonumber 
\end{equation}

Due to the nature of biopharmaceutical manufacturing, the state transition model $p(\pmb{s}_{t+1}|\pmb{s}_t,\pmb{a}_t;\pmb\theta)$ is highly complex, non-linear, and nonstationary. 
The state transition $ p(\pmb{s}_{t+1}|\pmb{s}_t, \pmb{a}_t;{\pmb\theta})$ is modeled by a hybrid (``mechanistic + statistical") model. 
Its structure takes existing mechanistic models as prior. For example, since the key factors influencing process dynamics and variability in the cell culture process are induced by cellular metabolisms \cite{o2021hybrid}, the probabilistic state transition of this KG hybrid model can incorporate 
cell metabolic/gene regulatory networks and account for cell-to-cell variations. 
\textit{Therefore, there are key properties in biomanufacturing process, specially for personalized cell/gene therapies, 
including (1) partially observed state ($\pmb{s}_t$) that means only limited proportion of state observable; (2) stochastic state transition model $p(\pmb{s}_{t+1}|\pmb{s}_t,\pmb{a}_t;\pmb\theta)$ involves high inherent stochasticity; 
and (3) very limited and heterogeneous online and offline measurement data.} 


Given limited historical observations, we focus on hybrid model Bayesian inference to support online mechanism learning, monitoring, and reliable interpretable prediction, accounting for both inherent stochasticity and model uncertainty.
The posterior distribution will be derived to quantify model uncertainty. 

\subsection{Hybrid Modeling for Bioprocess with Partially Observed State} 
\label{subsec:hybridModeling} 

At any time $t$, the process state is composed of observable and latent state variables, i.e., $\pmb{s}_t=(\pmb{x}_t,\pmb{z}_t)$ with  $\pmb{x}_t\in \mathcal{S}_x$ and latent variables $\pmb{z}_t\in \mathcal{S}_z$, 
where $\mathcal{S}_x\subset \mathbb{R}^{d_x}$ and $\mathcal{S}_z\subset \mathbb{R}^{d_z}$ with $\mathcal{S}=\mathcal{S}_x\times \mathcal{S}_z$ and $d=d_x+d_z$. 
Denote the partially observed state trajectory as $\pmb\tau_x\equiv (\pmb{x}_1,\pmb{a}_1,\ldots,\pmb{x}_H,\pmb{a}_H,\pmb{x}_{H+1})$. 
Given model parameters $\pmb{\theta}$, by integrating out latent variables $(\pmb{z}_1, \ldots,  \pmb{z}_{H+1})$, 
the likelihood evaluation of any observation $\pmb{\tau}_x$, i.e.,
\begin{equation}
    p(\pmb{\tau}_x| \pmb\theta)= \int
    \cdots \int 
    p(\pmb\tau|\pmb\theta) d \pmb{z}_1 \cdots d \pmb{z}_{H+1},
    \label{eq.likelihood}
    \nonumber 
\end{equation}
is intractable especially when the dimensions of model parameters and latent states are high.
This hybrid model characterizes the risk- and science-based understanding of underlying bioprocess mechanisms and quantifies spatial-temporal causal interdependencies of CPPs/CQAs. It can connect heterogeneous online and offline measures to infer unobservable state (such as metabolic state determining cell product functional behaviors and critical quality attributes), support process monitoring, and facilitate real-time release.

We model the bioprocess state transition with a hybrid (``mechanistic and statistical") model. Given the existing ODE-based mechanistic model, ${\mbox{d}\pmb{s}}/{\mbox{d}t} = \pmb{f}\left(\pmb{s},\pmb{a};\pmb\phi\right),$
by using the finite difference approximation for derivatives, i.e., $\mbox{d} \pmb{s}\approx \Delta \pmb{s}_t=\pmb{s}_{t+1}-\pmb{s}_t$, and $\mbox{d}t\approx \Delta t$, 
we construct the hybrid model for state transition, 
\begin{align}
    \pmb{x}_{t+1} = \pmb{x}_t + \Delta t \cdot \pmb{f}_x(\pmb{x}_t,\pmb{z}_t,\pmb{a}_t; \pmb\phi) + \pmb{e}^{x}_{t+1}
    ~~~ \mbox{and} ~~~
    \pmb{z}_{t+1} = \pmb{z}_t + \Delta t \cdot \pmb{f}_z(\pmb{x}_t,\pmb{z}_t,\pmb{a}_t; \pmb\phi) + \pmb{e}^{z}_{t+1},
    \nonumber 
\end{align}
with unknown 
kinetic coefficients $\pmb\phi\in \mathbb{R}^{d_{\phi}}$ (e.g., cell growth and inhibition rates). The function structures of $\pmb{f}_x(\cdot)$ and $\pmb{f}_z(\cdot)$ are the parts of $\pmb{f}(\cdot)$ associated to the observable state output $\pmb{x}_{t+1}$ and the latent state output $\pmb{z}_{t+1}$.
By applying the central limit theorem, the residual terms, accounting for inherent stochasticity and other factors, are modeled by  multivariate Gaussian distributions $\pmb{e}_{t+1}^{x} \sim \mathcal{N}(0,V^{x})$ and $\pmb{e}_{t+1}^{z} \sim \mathcal{N}(0,V^{z})$ with zero means and covariance matrices $V^{x}$ and $V^{z}$.
Then, the state transition distribution becomes,
\begin{align*}
     \pmb{x}_{t+1}|\pmb{x}_t,\pmb{z}_t,\pmb{a}_t  
    \sim \mathcal{N}\Big(\pmb{x}_t + \Delta t \cdot \pmb{f}_x(\pmb{x}_t,\pmb{z}_t,\pmb{a}_t),  V_{t+1}^{x} \Big)
    ~~~\mbox{and} ~~~
     \pmb{z}_{t+1}|\pmb{x}_t,\pmb{z}_t,\pmb{a}_t
    \sim  \mathcal{N}\Big(\pmb{z}_t + \Delta t \cdot \pmb{f}_z(\pmb{x}_t,\pmb{z}_t,\pmb{a}_t),  V_{t+1}^{z} \Big). 
\end{align*} 
Thus, the stochastic state transition model, specified by parameters $\pmb{\theta}=(\pmb\phi,V^{{x}},
V^{{z}})^\top$, characterizes the bioprocess inherent stochasticity, dynamics, and mechanisms (such as biophysicochemical reactions).


\subsection{Challenges of Hybrid Model Inference Under High Stochasticity and Limited Data}
\label{subsec:ModelInferenceChallenges}


Given limited real-world data with size $m$, denoted by $\mathcal{D}=\{\pmb\tau_x^{(i)}: i=1,2,\ldots,m\}$, the model uncertainty is quantified by a posterior distribution derived through applying the Bayes' rule,
\begin{equation}
p(\pmb{\theta}|\mathcal{D}) \propto p(\pmb{\theta})p(\mathcal{D}|\pmb{\theta})=p(\pmb{\theta})\prod_{i=1}^m p\left(\left.\pmb{\tau}_x^{(i)} \right|\pmb{\theta} \right),
\label{eq.posterior}
\end{equation}
where $p(\pmb{\theta})$ represents the prior distribution. 
It is challenging to directly derive or computationally assess the posterior distribution $p(\pmb{\theta}|\mathcal{D})$ in eq.~(\ref{eq.posterior}). 
First, there often exist large-dimensional latent state variables $\pmb{z}_t$, especially for multi-scale bioprocess model characterizing the scientific understanding of individual cell response to micro-environmental perturbation and accounting for cell-to-cell variation in metabolic/gene networks. It is computationally expensive to assess the likelihood for each observation, $p(\pmb{\tau}_x^{(i)}| \pmb\theta)= \int \cdots \int p(\pmb\tau^{(i)}|\pmb\theta) d \pmb{z}_1 \cdots d \pmb{z}_{H+1}$ with $i=1,2,\ldots,m$, especially for bioprocess with optical sensor online monitoring (that means the value of $H$ is large). Second, the mechanistic model $\pmb{f}(\pmb{s},\pmb{a}; \pmb\phi)$ can be a nonlinear function of state $\pmb{s}$ and parameters $\pmb\phi$. 
The random kinetic coefficients often have 
 batch-to-batch variations. 
For example, the kinetic coefficients (such as cell growth rate, oxygen/nutrient uptake rates, and metabolic waste excretion rate) can depend on the gene expression of seed cells and cell culture environments. 
Third, the amount of real-world process observations can be very limited (especially for personalized bio-drug manufacturing) even though inherent stochasticity and model uncertainty are high. 

Thus, in Section~\ref{sec:BayesianInference}, ABC approach is considered to approximate the posterior distribution of KG hybrid model with high fidelity that can capture the key features of biomanufacturing processes.
Since it is computationally expensive especially under the situations with high stochastic and model uncertainties, LG-DBN auxiliary ABC-SMC is created to 
facilitate the Bayesian inference. 
Based on Taylor series approximation of the hybrid model, this linear auxiliary model can be accurate for biomanufacturing process with optical sensor (e.g., fluorescent probe and Raman sensors) online monitoring.

\section{SEQUENTIAL IMPORTANCE SAMPLING FOR BAYESIAN INFERENCE} 
\label{sec:BayesianInference}

When the evaluation of likelihood for each observation is computationally intractable, i.e., $p(\pmb{\tau}_x^{(i)}| \pmb\theta)= \int \cdots \int p(\pmb\tau^{(i)}|\pmb\theta) d \pmb{z}_1 \cdots d \pmb{z}_{H+1}$ for $i=1,2,\ldots,m$, 
the ABC approach is recommended to approximate the posterior distribution \cite{sisson2018handbook}. 
In the naive ABC implementation, we draw a candidate sample from the prior $\pmb{\theta} \sim p(\pmb{\theta})$ and then generate a simulation dataset $\mathcal{D}^\star$ from the hybrid model. 
If the simulated dataset $\mathcal{D}^\star$ is ``close" to the observed real-world observations $\mathcal{D}$, we accept the sample $\pmb{\theta}$; otherwise reject it. 
Thus, we approximate the posterior distribution $p(\pmb{\theta}|\mathcal{D})$
with $p(\pmb{\theta}|d\left(\mathcal{D},\mathcal{D}^\star\right) \leq h)$, where 
$d(\cdot)$ is a distance metric (e.g., Euclidean distance, likelihood distance) and $h$ is an approximation tolerance level.


However, for any given small tolerance level $h$, we often face very low accept rate for complex biomanufacturing processes with high stochastic and model uncertainties.
The random discrepancy between process trajectories $\mathcal{D}$ and $\mathcal{D}^\star$ can be large even when the parameter sample $\pmb{\theta}$ equals to $\pmb{\theta}^c$. In addition, given very limited real-world data for the complex hybrid model, the dimension of model parameters $\pmb{\theta}$ is large and the model uncertainty can be high.

To increase the accept rate and ensure the computational efficient generation of  samples $\pmb{\theta}$ with good approximation on the critical features occurring in the real-world data, we will design the distance measure $d(\cdot)$ based on \textit{designed} lower dimensional summary statistics, denoted by ${\eta}(\mathcal{D})$, in Section~\ref{sec:summarystats}. 
It means that we accept samples $\pmb{\theta}$ which lead to the summary statistics of simulated data, denoted by $\eta^\star = \eta(\mathcal{D}^\star)$, close to that of observations $\eta_{obs}=\eta(\mathcal{D})$. 
Thus, the standard ABC framework \cite{sisson2018handbook} becomes 
\begin{equation}
    p_{ABC}(\pmb{\theta}|\eta_{obs}) \propto \int 
    \mathbbm{1}(d(\eta^\star,\eta_{obs}) \leq h)
    p(\eta^\star|\pmb{\theta})p(\pmb{\theta})d\eta^\star.
    \label{eq.ABC_summaryStatistics}
\end{equation}
As the distance tolerance $h$ gradually decreases, we have
\begin{equation}
    \lim_{h \rightarrow 0} p_{ABC}(\pmb{\theta}|\eta_{obs}) \propto \int \delta_{\eta_{obs}}(\eta^\star) p(\eta^\star|\pmb{\theta})p(\pmb{\theta})d\eta^\star = p(\eta_{obs}|\pmb{\theta})p(\pmb{\theta}) \propto p(\pmb{\theta}|\eta_{obs}),
    \nonumber
\end{equation}
where $\delta_X(x)$ denotes the Dirac measure, defined as $\delta_X(x) = 1$ if $x = X$ and $\delta_X(x) = 0$ otherwise.

\textit{A good design of ABC summary statistics $\eta$ should balance complexity 
v.s. informativeness.} If the summary statistics $\eta$ are sufficient for $\pmb{\theta}$, then $p(\pmb{\theta}|\eta_{obs})$ will be equivalent to $p(\pmb{\theta}|\mathcal{D})$. With small threshold $h$, the ABC approximate $p_{ABC}(\pmb{\theta}|\eta_{obs})$ in (\ref{eq.ABC_summaryStatistics}) can provide a good approximation of the true posterior. However, in the most situations, it is challenging to specify sufficient statistics since the KG hybrid model is built based on mechanistic models and it accounts for the key features including (1) partially observed state; (2) heterogeneous offline and online measures; (3) nonlinear mechanisms and dynamics; and (4) batch-to-batch variations on mechanistic coefficients.
\textit{Therefore, in Section~\ref{sec:summarystats}, we project the bioprocess KG hybrid model into linear Gaussian dynamic Bayesian Network (LG-DBN) auxiliary model space that has tractable likelihood. It can capture first two moments of bioprocess dynamics and variations to support robust and optimal control}. We will use the LG-DBN likelihood to derive summary statistics accelerating the generation of critical samples $\pmb{\theta}$. 
Our study also shows that complex KG hybrid models will asymptotically converge to a LG-DBN model as time interval $\Delta t$ becomes ``smaller and smaller" by applying Taylor approximation \cite{zheng2021policy}. 
Thus, this LG-DBN approximation holds well for many cases with online sensor monitoring and bioprocess (e.g.,  biological state of cells) that does not change quickly.

The basic ABC generates candidate samples from the prior $p(\pmb{\theta})$ and uses the accept/reject approach to retain those samples satisfying the approximation threshold requirement.
This can be extremely ineffective especially for the situations using noninformative prior that has a wide sampling space. 
The \textit{ABC-sequential Monte Carlo (ABC-SMC) methods} derived from the sequential importance sampling 
\cite{toni2009approximate,beaumont2009adaptive} can improve the sampling efficiency through generating candidate samples from updated posterior approximates. 
In specific, let $g$ denote the index of ABC iterations used to improve the approximation of the posterior distribution $p(\pmb{\theta}|\mathcal{D})$.
We select a sequence of intermediate target distribution, 
denoted by $\{\pi_g\}$ for $g = 1,2, \ldots, G$, converging to $p(\pmb{\theta}|\mathcal{D})$ as
we gradually reduce the tolerance level $h_g$, 
\begin{equation}
    \pi_g(\pmb{\theta}) = p(\pmb{\theta})\mathbbm{1}
    \left(d(\eta^\star,\eta_{obs}) \leq h_g \right).
    \label{eq.intermediaTargetDistribution}
\end{equation}
Direct sampling from $p(\pmb{\theta})$ and having the accept/reject based on the condition $\mathbbm{1}
    \left(d(\eta^\star,\eta_{obs}) \leq h_g \right)$ in (\ref{eq.intermediaTargetDistribution}) is not simulation efficient. The accept rate can be low as $h_g$ becomes smaller and smaller.
    
Thus, we use the \textit{sequential importance sampling (SIS)} and select a sequence of proposal distribution, denoted by $\{\zeta_g \}$ for $g=1,2,\ldots,G$, to improve the sampling efficiency, i.e.,
\begin{equation}
    \zeta_{g}(\pmb{\theta}) = \mathbbm{1}\left(\pi_g(\pmb{\theta}) >0 \right)\int \pi_{g-1}(\pmb{\theta}^\prime)K(\pmb{\theta}^\prime,\pmb{\theta}) d \pmb{\theta}^\prime,
    \label{eq.intermediaProposalDistribution}
\end{equation}
where $K(\pmb{\theta}^\prime,\pmb{\theta})$ is  a Markov kernel.
The proposal distribution $\zeta_g(\pmb{\theta})$ is defined as the perturbed previous intermediate
distribution $\pi_{g-1}$ through the perturbation kernel $K$.
After generating $N$ sample particles from the proposal distribution $\pmb{\theta}_n \sim \zeta_g(\pmb{\theta})$ for $n=1,2,\ldots,N$, we weight it by $w_n^{(g)} = {\pi_g}(\pmb{\theta}_n)/{\zeta_g}(\pmb{\theta}_n)$. The condition, $\mathbbm{1}(\pi_g(\pmb{\theta}) >0)$, in (\ref{eq.intermediaProposalDistribution}) is used to satisfy the importance sampling condition, i.e., 
$\{\pmb{\theta}: \pi_g (\pmb{\theta}) > 0 \} \subset \{ \pmb{\theta}: \zeta_g (\pmb{\theta}) > 0\}$.  
This can avoid the weight becoming infinite, which will lead to high variance on the SIS estimator.
We set the first proposal distribution to be the prior distribution, i.e., $\zeta_1(\pmb{\theta}) = p(\pmb{\theta})$.

\begin{algorithm}[th] 
\DontPrintSemicolon
\KwIn{
the prior distribution $p(\pmb{\theta})$; the number of particles $N$; 
process observations $\mathcal{D} = \{\pmb\tau_x^{(i)}\}_{i=1}^m$;  the perturbation kernel function $K(\cdot)$; the number of particles to keep at each iteration 
$N_\alpha = \lfloor \alpha N \rfloor$ with $\alpha \in [0,1]$; and the minimal acceptance rate $p_{acc_{min}}$.
}
\KwOut{posterior distribution approximate $\widehat{p}(\pmb{\theta}|\mathcal{D})$.
 } 
{   
    \For{$n = 1,\ldots, N$}{
    \textbf{1.} Sample $\pmb{\theta}^{(0)}_n\sim p(\pmb{\theta})$; \\
    \textbf{2.} 
    Generate $m \times L$ predicted trajectories $\mathcal{D}^\star= \{\pmb\tau_{x}^{\star(i)}\}_{i=1}^{mL}$ using $\pmb{\theta}^{(0)}_n$; \\ 
    \textbf{3.} Set $q_n^{(0)} =  d(\eta(\mathcal{D}), \eta(\mathcal{D}^\star))$ and $w_n^{(0)} = 1$;
    }
    \textbf{4.} Let $h_1$ 
    be the first $\alpha$-quantile of $q^{(0)} = \{q_n^{(0)}\}_{n = 1}^{N}$;\\
    \textbf{5.} Let $\{(\pmb{\theta}_n^{(1)},w_n^{(1)},q_n^{(1)})\} = \{(\pmb{\theta}_n^{(0)},w_n^{(0)},q_n^{(0)})|q_n^{(0)} \leq h_1, 1\leq n \leq N\}$, $p_{acc} = 1$ and $g = 2$;\\
    \While{$p_{acc} > p_{acc_{min}}$}{
        \For{$n = N_\alpha + 1,\ldots,N$}{
            \textbf{6.} Sample $\pmb{\theta}^\star_n$ from $\pmb\theta^{(g-1)}_k$ with probability $\frac{w^{(g-1)}_k}{\sum_{j=1}^{N_\alpha}w_j^{(g-1)}}$, $1 \leq k \leq N_\alpha$;\\
            \textbf{7.} Perturb the particle to obtain $\pmb{\theta}^{(g-1)}_n \sim K(\pmb{\theta}|\pmb{\theta}^{\star}_n) = \mathcal{N}(\pmb{\theta}^{\star}_n,\sum)$; \\
            \textbf{8.} 
            Generate $m \times L$ predicted trajectories $\mathcal{D}^\star = \{\pmb\tau_{x}^{\star(i)}\}_{i=1}^{mL}$ using $\pmb{\theta}^{(g-1)}_n$: \\ 
            \textbf{9.} Set $q_n^{(g-1)} =  d(\eta(\mathcal{D}), \eta(\mathcal{D}^\star))$;\\
            \textbf{10.} Set ${w}_n^{(g-1)} =   \frac{p(\pmb{\theta}^{(g-1)}_n) \mathbbm{1}(d(\eta(\mathcal{D}), \eta(\mathcal{D}^\star)) \leq h_{g-1})}{\sum_{j=1}^{N_\alpha} \frac{w_j^{(g-1)}}{ \sum_{k=1}^{N_\alpha} w_k^{(g-1)}} K(\pmb{\theta}_n^{(g-1)}|\pmb{\theta}^{(g-1)}_{j})}$;
        }
    \textbf{11.} Set $p_{acc} = \frac{1}{N-N_\alpha}\sum_{k=N_\alpha+1}^N \mathbbm{1}(q_k^{(g-1)} \leq h_{g-1})$; \\
    \textbf{12.} Let $h_g$ 
    be the first $\alpha$-quantile of 
    $q^{(g-1)} = \{q_n^{(g-1)}\}_{n = 1}^{N}$;\\
    \textbf{13.} Let $\{(\pmb{\theta}_n^{(g)},w_n^{(g)},q_n^{(g)})\} = \{(\pmb{\theta}_n^{(g-1)},w_n^{(g-1)},q_n^{(g-1)})|q_n^{(g-1)} \leq h_{g}, 1\leq n \leq N\}$ and $g = g + 1$;\\
    }
    \textbf{14. Return} the approximated posterior distribution, $\widehat{p}(\pmb{\theta}|\mathcal{D}) = \frac{1}{\sum_{n'=1}^{N_\alpha} w^{(g-1)}_{n'}} \sum_{n=1}^{N_\alpha} w^{(g-1)}_n
    \delta_{\pmb{\theta}_n^{(g-1)}}(\pmb{\theta})$.
}
\caption{DBN auxiliary based ABC-SMC for hybrid model Bayesian inference. 
}
\label{Algr:SMC-ABC}
\end{algorithm}

\begin{sloppypar}
\textit{The proposed LG-DBN auxiliary likelihood-based ABC-SMC sampling procedure is summarized in Algorithm~\ref{Algr:SMC-ABC}}.
It incorporates an adaptive selection approach on the threshold $h_g$ from \cite{toni2009approximate,lenormand2013adaptive,del2006sequential}. The initial set of parameter samples $\{\pmb{{\theta}}_{n}^{(0)}\}_{n=1}^N$ is generated from the prior distribution $p({\pmb{\theta}})$ in {Step~1}. The associated weights $\{w_{n}^{(0)}\}_{n=1}^N$ and distances $\{q_{n}^{(0)}\}_{n=1}^N$ are calculated in {Steps~2-3}. 
Considering the impact from stochastic uncertainty, we generate $m L$ predicted trajectories denoted by $\mathcal{D}^\star= \{\pmb\tau_{x}^{\star(i)}\}_{i=1}^{mL}$, compute the LG-DBN auxiliary based summary statistics $\eta(\mathcal{D}^\star)$, and then calculate the distance $q_n^{(0)}$. 
The tolerance level $h_g$ in any $g$-th iteration is determined online as the $\alpha$-quantile of the $\{q_n^{(g)}\}_{n = 1}^{N}$. The particles, satisfying this tolerance denoted by $\{\pmb{\theta}_n\}_{n=1}^{N_\alpha}$, constitute the weighted empirical distribution 
to approximate the posterior distribution in {Steps~5 and 13}, where $N_\alpha = \lfloor \alpha N \rfloor$. 
The approximation accuracy is measured by the corresponding distances $\{q_{n}\}_{n=1}^{N_\alpha}$. 
Then, $N-N_\alpha$ new particles are drawn from the proposal distribution $\zeta_{g}(\pmb{\theta})$ 
in {Steps~6-7}. 
The associated weights and distances are calculated in {Steps~8-10}. The tolerance level $h_g$ and the posterior distribution approximate $\pi_g(\pmb{\theta})$ are updated in {Steps 12-13}. We repeat Steps~6-13 until the proportion 
of particles satisfying the tolerance level $h_{g-1}$ among the $N-N_\alpha$ new particles is below the pre-specified threshold $p_{acc_{min}}$. Finally, the ABC-SMC algorithm returns the weighted empirical distribution, denoted by $\widehat{p}(\pmb{\theta}|\mathcal{D})$, as posterior distribution approximate in Step~14.
\end{sloppypar}



\section{LG-DBN AUXILIARY LIKELIHOOD-BASED SUMMARY STATISTICS}
\label{sec:summarystats}


Motivated by the studies  \cite{martin2019auxiliary,gleim2013approximate}, in this section, we derive LG-DBN auxiliary likelihood-based summary statistics for ABC-SMC to capture the crucial features of the bioprocess trajectory on dynamics and variations. 
Given a set of observations $\mathcal{D}=\{\pmb\tau_x^{(i)}: i=1,2,\ldots,m\}$, 
we derive the MLE of LG-DBN auxiliary model, i.e., maximizing the log-likelihood $\hat{\pmb\beta}(\mathcal{D}) = \mbox{argmax}_{\pmb{\beta}}\ell(\pmb{\beta}|\mathcal{D})$. Then we use it as the summary statistics $\eta\triangleq\hat{\pmb\beta}$ to calculate the distance measure $q \equiv d(\hat{\pmb\beta},\hat{\pmb\beta}^\star)$, where $\hat{\pmb\beta}^\star$ is the summary statistics of simulated data. In the following, we first develop the LG-DBN model with only observable state transition in Section~\ref{subsec:LinearAuxiliaryModel} and then discuss the parameter estimation in Section~\ref{subsec:summaryStatistics}.
\vspace{-0.1in}

\subsection{The development of LG-DBN Auxiliary Model}
\label{subsec:LinearAuxiliaryModel}

\begin{sloppypar}
Let $x_{1}^k \sim \mathcal{N}(\mu^{x,k}_{1},(v^{x,k}_{1})^2)$ with $k=1,2\ldots,d$
model the variation in the $k$-th initial observed state. 
In practice, to ensure product quality, CPPs are strictly regulated by the specifications of ranges of values. Thus, we model $\pmb{a}_t$ as a random variable, i.e., $a_{t}^k \sim \mathcal{N}(\lambda^{x,k}_{t},(\sigma^{x,k}_{t})^2)$ with $k=1,2\ldots,d_a$ and $t=1,2\ldots,H$. 
At any time $t$, 
the LG-DBN auxiliary model has the state transition model,
\vspace{-0.1in}

\begin{equation}
    \pmb{x}_{t+1}= \pmb{\mu}^x_{t+1} + \pmb{\psi}_t^{x}  (\pmb{x}_t-\pmb{\mu}^x_{t}) + \pmb{\psi}^{a}_t (\pmb{a}_t-\pmb{\mu}^a_{t}) + (V_{t+1}^{x})^{\frac{1}{2}} \pmb{\omega},
    \label{eq: linearBN}
    \vspace{-0.05in}
\end{equation}

\noindent where $\pmb\mu_{t}^x=(\mu_t^{1},\ldots, \mu_t^{d_x})$, $\pmb\mu_{t}^a=(\lambda_t^{1},\ldots, \lambda_t^{d_a})$, $\pmb{\omega}$ is an $d_x$-dimensional standard normal random vector, and $V_{t+1}^x=\mbox{diag}
((v_{t+1}^{x,k})^2)$ is a diagonal covariance matrix. 
The coefficients $\pmb{\psi}^{{x}}_t$ and $\pmb{\psi}^{{a}}_t$ measure the main effects of current observed state $\pmb{x}_t$ and action $\pmb{a}_t$ on the next observed state $\pmb{x}_{t+1}$.  
Let $\pmb\sigma_t=(\sigma_t^1,\ldots,\sigma_t^{d_a})$ and $\pmb{v}^x_t=(v_t^{x,1},\ldots,v_t^{x,d_x})$. Thus, the LG-DBN model, specified by parameters  $\pmb{\beta} = (\pmb{\mu}^x,\pmb\mu^a,\pmb{\psi}^x,\pmb{\psi}^a,\pmb\sigma,\pmb{v}^x)= \{(\pmb{\mu}_{t}^x,\pmb\mu_t^a,\pmb{\psi}_{t}^x,\pmb\psi_{t}^a,\pmb\sigma_t,\pmb{v}^x_t)| 1\leq t\leq H\}$, has the joint distribution of bioprocess trajectory: $p(\pmb{\tau}_x) = p(\pmb{x}_1,\pmb{a}_1,\ldots,\pmb{x}_H,\pmb{a}_H,\pmb{x}_{H+1})      = p(\pmb{x}_1) \prod_{t=1}^H p(\pmb{x}_{t+1}|\pmb{x}_{t},\pmb{a}_t)p(\pmb{a}_t)$.
\end{sloppypar}

\begin{sloppypar}
Let $\pmb{\mu}_\tau = [\pmb{\mu}^x_1,\pmb{\mu}^a_1,\ldots,\pmb{\mu}^x_{H},\pmb{\mu}^a_{H},\pmb{\mu}^x_{H+1}]^\top$. Following \cite{murphy2012machine},  we rewrite \eqref{eq: linearBN} in the following form
\vspace{-0.1in}
\begin{equation}
    \pmb{\tau}_x-\pmb\mu_\tau=B(\pmb{\tau}_x - \pmb\mu_\tau)+ \Sigma^{\frac{1}{2}}_\tau \pmb{\omega}_\tau \label{eq: general matrix form of linear Observable BN}
\end{equation}
where $\pmb{\omega}_\tau$ is an $((H+1)d_x+Hd_a)$-dimensional standard normal random vector, $\Sigma^{\frac{1}{2}}_\tau=\text{diag}
(\pmb{v}^x_1,\pmb\sigma_1,\ldots,\pmb{v}^x_H,\pmb\sigma_H,\pmb{v}^x_{H+1})$ is the diagonal matrix of the conditional standard deviations of observed state and actions, and the coefficient matrix of observed trajectory is written as

\begin{center} \small
\setcounter{MaxMatrixCols}{20}
$B = \begin{bmatrix} 
0 & 0 & 0 & 0 & 0 & 0 & \cdots & 0 & 0 & 0 & 0 \\
0 & 0 & 0 & 0 & 0 & 0 & \cdots & 0 & 0 & 0 & 0\\
\pmb{\psi}^x_1 & \pmb{\psi}^a_1 & 0 & 0 & 0 & 0 & \cdots & 0 & 0 & 0 & 0\\
0 & 0 & 0 & 0 & 0 & 0 & \cdots & 0 & 0 & 0 & 0\\
0 & 0& \pmb{\psi}^x_2 & \pmb{\psi}^a_2 &  0 & 0 &\cdots & 0 & 0 & 0 & 0\\
0 & 0 & 0 & 0 & 0 & 0 & \cdots & 0 & 0 & 0 & 0\\
\vdots & \vdots & \vdots & \vdots & \vdots & \vdots & \vdots & \vdots & \vdots & \vdots & \vdots\\ 
0 & 0 & 0 & 0 & 0 & 0 & \cdots & \pmb{\psi}^x_H & \pmb{\psi}^a_H & 0 & 0\\
\end{bmatrix}$.
\end{center}
Thus, by rearranging \eqref{eq: general matrix form of linear Observable BN} and letting $\pmb{\tau}_x-\pmb\mu_\tau=(I-B)^{-1}\Sigma_\tau^{\frac{1}{2}} \pmb{\omega}_\tau$, 
 we have $\pmb{\tau}_x\sim \mathcal{N}(\pmb\mu_\tau, (I-B)^{-1}\Sigma_\tau (I-B)^{-\top})$ 
with mean $\E[\pmb\tau_x]=\pmb{\mu}_\tau$ and covariance matrix 
$\mbox{Cov}(\pmb{\tau}_x-\pmb\mu_\tau)=(I-B)^{-1}\Sigma_\tau (I-B)^{-\top}$. 
\end{sloppypar}

\subsection{Linear Gaussian Dynamic Bayesian Network based Summary Statistics}
\label{subsec:summaryStatistics}

Let $\tilde{\pmb\tau}_x 
\equiv (\tilde{\pmb{x}}_1,\tilde{\pmb{a}}_1,\ldots,\tilde{\pmb{x}}_H,\tilde{\pmb{a}}_H,
\tilde{\pmb{x}}_{H+1})
=\pmb{\tau}_x - \pmb\mu_\tau$, where $\tilde{\pmb{x}}_t$ and $\tilde{\pmb{a}}_t$ denote centered observable state and decision.
Given $m$ observations $\mathcal{D} = \{\pmb{\tau}_x^{(i)}\}_{i=1}^m$, 
the unbiased estimator $\hat{\pmb\mu}_\tau=\frac{1}{m}\sum^m_{i=1}\pmb\tau^{(i)}_x$ can be easily obtained by using the fact $\E[\pmb\tau_x]=\pmb\mu_\tau$.
The log-likelihood of the centered trajectory observations $\{\tilde{\pmb{\tau}}^{(i)}_x\}_{i=1}^m$ becomes, 
\vspace{-0.05in}
\begin{align}
\max_{\pmb{{\psi}}^x, \pmb{{\psi}}^a,V} & \ell\left(\Tilde{\pmb\tau}_x^{(1)},\ldots, \Tilde{\pmb\tau}_x^{(m)}; \pmb{{\psi}}^x, \pmb{{\psi}}^a,V \right) = \max_{\pmb{{\psi}}^x, \pmb{{\psi}}^a,V}\log\prod_{i=1}^m p\left(\tilde{\pmb\tau}_x^{(i)}\right) \nonumber\\
  &=\max_{V_1} \sum_{i=1}^m\log p(\tilde{\pmb{x}}_1^{(i)}) \left[\sum_{t=1}^H\max_{\sigma_t}\sum_{i=1}^m \log p(\tilde{\pmb{a}}_t^{(i)}) \right] \left[\sum_{t=1}^H\max_{\pmb{{\psi}}_t^x,\pmb{{\psi}}_t^a,\pmb{v}^x_{t+1}} \sum_{i=1}^m\log p(\tilde{\pmb{x}}_{t+1}^{(i)}|\tilde{\pmb{x}}_t^{(i)},\tilde{\pmb{a}}_t^{(i)})\right].
  \nonumber
\end{align}
Since both initial state $\tilde{\pmb{x}}_1$ and actions $\tilde{\pmb{a}}_t$ for $t=1,\ldots, H$ are normally distributed with mean zero, the MLEs of their variance are sample covariances: $\hat{v}^{x,k}_1=\frac{1}{m}\sum^m_{i=1}
(\tilde{{x}}_1^{k{(i)}})^2$ with $k=1,2,\ldots,d_x$ and $\hat{\sigma}_t^k=\frac{1}{m}\sum^m_{i=1}
(\tilde{{a}}_t^{k{(i)}})^2$ with $k=1,2,\ldots,d_a$. 
In addition, at any time $t$,  we have the log-likelihood of a sample $\Tilde{\pmb\tau}_x^{(i)}$
\begin{equation}
   \log p(\tilde{\pmb{x}}_{t+1}^{(i)}|\tilde{\pmb{x}}_t^{(i)},\tilde{\pmb{a}}_t^{(i)}) \propto
   - \frac{m}{2} \log|V_{t+1}^x|-\frac{1}{2} \left(\tilde{\pmb{x}}^{(i)}_{t+1}-\pmb{{\psi}}_t^x\tilde{\pmb{x}}^{(i)}_{t} -\pmb{{\psi}}_t^a\tilde{\pmb{a}}^{(i)}_{t}\right)^\top V^x_{t+1}\left(\tilde{\pmb{x}}^{(i)}_{t+1}-\pmb{{\psi}}_t^x\tilde{\pmb{x}}^{(i)}_{t} -\pmb{{\psi}}_t^a\tilde{\pmb{a}}^{(i)}_{t}\right).
   \nonumber
\end{equation}
Let $\tilde{\pmb{x}}_{t+1}^{(i)}$ and $(\tilde{\pmb{x}}_{t}^{(i)}, \tilde{\pmb{a}}_{t}^{(i)})$ denote the $i$-th rows of output matrix $Y$ and input matrix $X$. 
Let $B_{t}= \left(\pmb{{\psi}}^x_t,\pmb{{\psi}}^a_t\right)^\top$ denote the coefficient vector. As a result, 
the MLEs of $\pmb{{\psi}}_t^x$ and $\pmb{{\psi}}_t^a$ are
\begin{equation}
   \left(\hat{\pmb\psi}_t^x,\hat{\pmb\psi}_t^a\right)^\top= \hat{B}_{t} = \arg\max_{B_{t}}-\frac{1}{2} \left(Y-XB_{t}\right)^\top (V^x_{t+1})^{-1}\left(Y-X B_{t}\right) = (X^\top (V^x_{t+1})^{-1}X)^{-1}X^\top (V^x_{t+1})^{-1} Y.
   \nonumber
\end{equation}
The MLE of each standard deviation can be computed by $\hat{v}^{x,k}_t=\sqrt{\frac{1}{m}\sum^m_{i=1}\left(\tilde{x}_t^{k{(i)}}\right)^2}$ \cite{fuller1978estimation}. In sum, given observations $\mathcal{D}$, the MLE of LG-DBN auxiliary model is $\hat{\pmb{\beta}} = (\hat{\pmb{\mu}}^x,\hat{\pmb\mu}^a,\hat{\pmb{\psi}}^x,\hat{\pmb{\psi}}^a,\hat{\pmb\sigma},\hat{\pmb{v}}^x)$.

\section{EMPIRICAL STUDY}
\label{sec:empiricalStudy}
In this section, we use the erythroblast cell therapy manufacturing example presented in \cite{glen2018mechanistic} to assess the performance of the proposed LG-DBN auxiliary likelihood-based ABC-SMC approach. 
\vspace{-0.1in}

\subsection{Hybrid Modeling for Cell Therapy Manufacturing Process}

The cell culture process of erythroblast exhibits two phases: a relatively uninhibited growth phase followed by an inhibited phase. 
\cite{glen2018mechanistic} identified that this reversible inhibition is caused by an unknown cell-driven factor rather than commonly known mass transfer or metabolic limitations. They developed an ODE-based mechanistic model describing the dynamics of an unidentified autocrine growth inhibitor accumulation 
and its impact on the erythroblast cell production process, i.e.,
\vspace{-0.15 in}

\begin{align}
    \frac{\diff\rho_t}{\diff t} = r_g \rho_t \Bigg (1 - \Big(1+e^{k_s(k_c-I_t)} \Big) ^{-1} \Bigg )
    \nonumber 
    ~~~ \mbox{and} ~~~
    \frac{\diff I_t}{\diff t} = \frac{\diff\rho_t}{\diff t} - r_d I_t,
    \nonumber 
\end{align}
where $\rho_t$ and $I_t$ represent the cell density and the inhibitor concentration (i.e., latent state) at time $t$.
The kinetic coefficients $\pmb\phi=\{r_g, k_s, k_c,r_d\}$ denote the cell growth rate, the inhibitor sensitivity, the inhibitor threshold, and the inhibitor decay. 
Then, we construct the hybrid model, i.e.,
\vspace{-0.1 in}
\begin{equation}
    \rho_{t+1} = \rho_t + \Delta t \cdot r_g \rho_t \Bigg (1 - \Big(1+e^{k_s(k_c-I_t)} \Big) ^{-1} \Bigg ) + e^{\rho}_{t} 
    ~~\mbox{and} ~~ 
    I_{t+1} = I_t + \Delta t \cdot \Bigg (\frac{\rho_{t+1}-\rho_t}{\Delta t} - r_d I_t \Bigg) + e^I_t, 
    \label{equ:hybridinhibition}
\end{equation}
where the residuals follow the normal distributions $e_{t}^{\rho} \sim \mathcal{N}(0,v_{\rho}^{2})$ and $e_{t}^{I} \sim \mathcal{N}(0,v_{I}^{2})$ by applying CLT. Therefore, the hybrid model is specified by parameters $\pmb\theta=(r_g, k_s, k_c, r_d, v_\rho, v_I)$. The prediction is made on the interval of three hours $\Delta t=3$ from 0 to 30 hours (corresponding to time step $t=1,2,\ldots, 11$). 

We denote the ``true" hybrid model with underlying parameters $\pmb\theta^c$. Following \cite{glen2018mechanistic}, we specify the true mechanistic parameter values as $\pmb\phi^c=\{r_g, k_s, k_c,r_d\} = \{0.057,3.4,2.6,0.005\}$. 
We set the bioprocess noise level $v = v_\rho = v_I $, 
 the initial cell density 3 $\times 10 ^6$ cells/mL (i.e., $\rho_1=3$), and no initial inhibition (i.e., $I_1=0$). Based on the simulation data generated by the true hybrid model, we assess the performance of the proposed LG-DBN auxiliary ABC-SMC algorithm under different levels of bioprocess noise $v = \{0.1,0.2\}$ and model uncertainty induced with the different data size, i.e., $m = 3, 6, 20$ batches.

\subsection{LG-DBN Auxiliary Sequential Importance Sampling Performance Assessment}
\label{subsec: performance}

We compare the performance of LG-DBN auxiliary ABC-SMC with naive ABC-SMC 
in terms of: (1) prediction accuracy, (2) computation time, and (3) posterior concentration. The distance metrics of naive ABC-SMC is $d\left(\mathcal{D},\mathcal{D}^\star\right)$. The results are estimated based 30 macro-replications. 
We set the number of particles $N = 400$, the ratio $\alpha = 0.5$, the number of replications $L = 60$, 
and the minimal accept rate $P_{acc_{min}} = 0.15$. 
The prior distributions of model parameters are set as: $r_g \sim U(0,0.5)$, $k_s \sim U(0,5)$, $k_c \sim U(0,5)$, $r_d \sim U(0,0.05)$, $v_\rho \sim U(0,0.2)$, and $v_I \sim U(0,0.2)$.

One of the major benefits induced by the LG-DBN auxiliary likelihood is that it provides an efficient way to measure the distance between simulated and observed samples, which quickly leads to posterior samples fitting well on dynamics and variations. To show the advantage of LG-DBN auxiliary  ABC-SMC, we first study its computational efficiency.
For each $r$-th macro replication, let $T_{w}^{(r)}$ and $T_{wo}^{(r)}$ represent the computation cost of the ABC-SMC algorithm with and without LG-DBN auxiliary. 
The computational efficiency improvement 
is evaluated as the time consuming ratio defined as $C^{(r)} = {T_{wo}^{(r)}}/{T_{w}^{(r)}}$. 
We record the 95\% confidence interval (CI) for improvement, denoted by $\bar{C} \pm 1.96 \times S_C/\sqrt{30}$, where $\bar{C} = \frac{1}{30} \sum_{r=1}^{30} {T_{wo}^{(r)}}/{T_{w}^{(r)}}$ and $S_C = [\sum_{r=1}^{30} ({T_{wo}^{(r)}}/{T_{w}^{(r)}} - \bar{C})^2/29]^{1/2}$; see the results in Table~\ref{table: timeimprove}.
With the LG-DBN auxiliary, the ABC-SMC algorithm shows significant improvement in computational efficiency. In all different settings, the mean computation cost of naive ABC-SMC 
is higher than the LG-DBN auxiliary based ABC-SMC by 27\% (at low variance and small sample size) to 163\% (at high variance and relative larger sample size).

\begin{wraptable}{r}{9 cm}
\vspace{-0.1in}
\small
\caption{Computational efficiency improvement ratio.}
\vspace{-0.05in}
\label{table: timeimprove}
\centering
\begin{tabular}{c|c|c|c}
\toprule
Process Noise & \multicolumn{1}{c|}{$m$ = 3}      &  \multicolumn{1}{c|}{$m$ = 6} &  $m$ = 20    \\ \midrule
$v = 0.1$       & \multicolumn{1}{c|}{1.27 $\pm$ 0.11}     &  \multicolumn{1}{c|}{1.43 $\pm$ 0.11}   & \multicolumn{1}{c}{2.44 $\pm$ 0.15}       \\ \midrule
$v = 0.2$      & \multicolumn{1}{c|}{1.39 $\pm$ 0.08}     &  \multicolumn{1}{c|}{1.52 $\pm$ 0.17}  & \multicolumn{1}{c}{2.63 $\pm$ 0.20}   \\ \bottomrule
\end{tabular}
\vspace{-0.05in}
\end{wraptable}

\begin{figure}[!thb]
     \centering
     \subfloat[Cell density (with LG-DBN auxiliary)]{%
         \centering
     \includegraphics[width=0.49\textwidth]{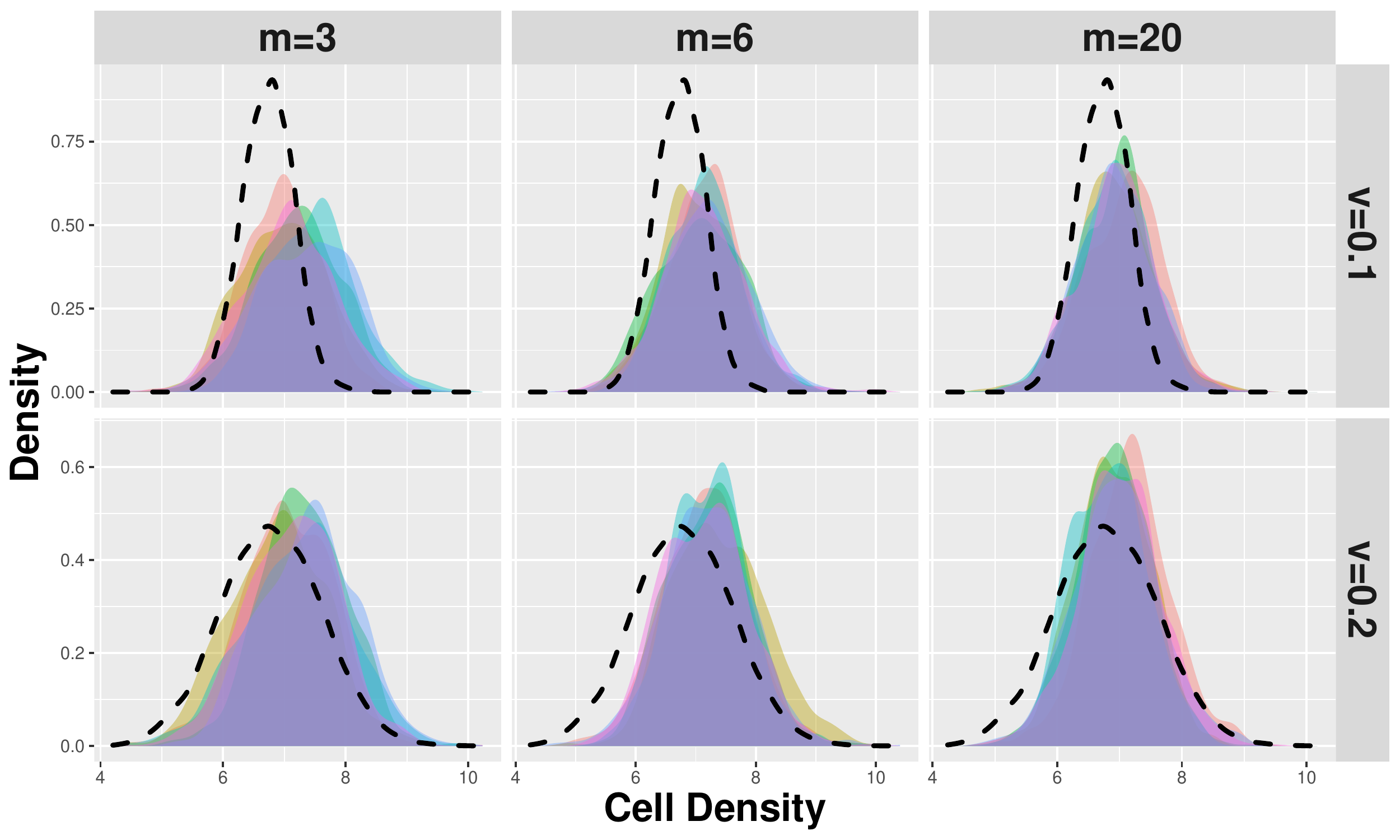}
         \label{fig: cell density with auxilary}}
     \subfloat[Inhibitor concentration (with LG-DBN auxiliary)]{
         \centering
     \includegraphics[width=0.49\textwidth]{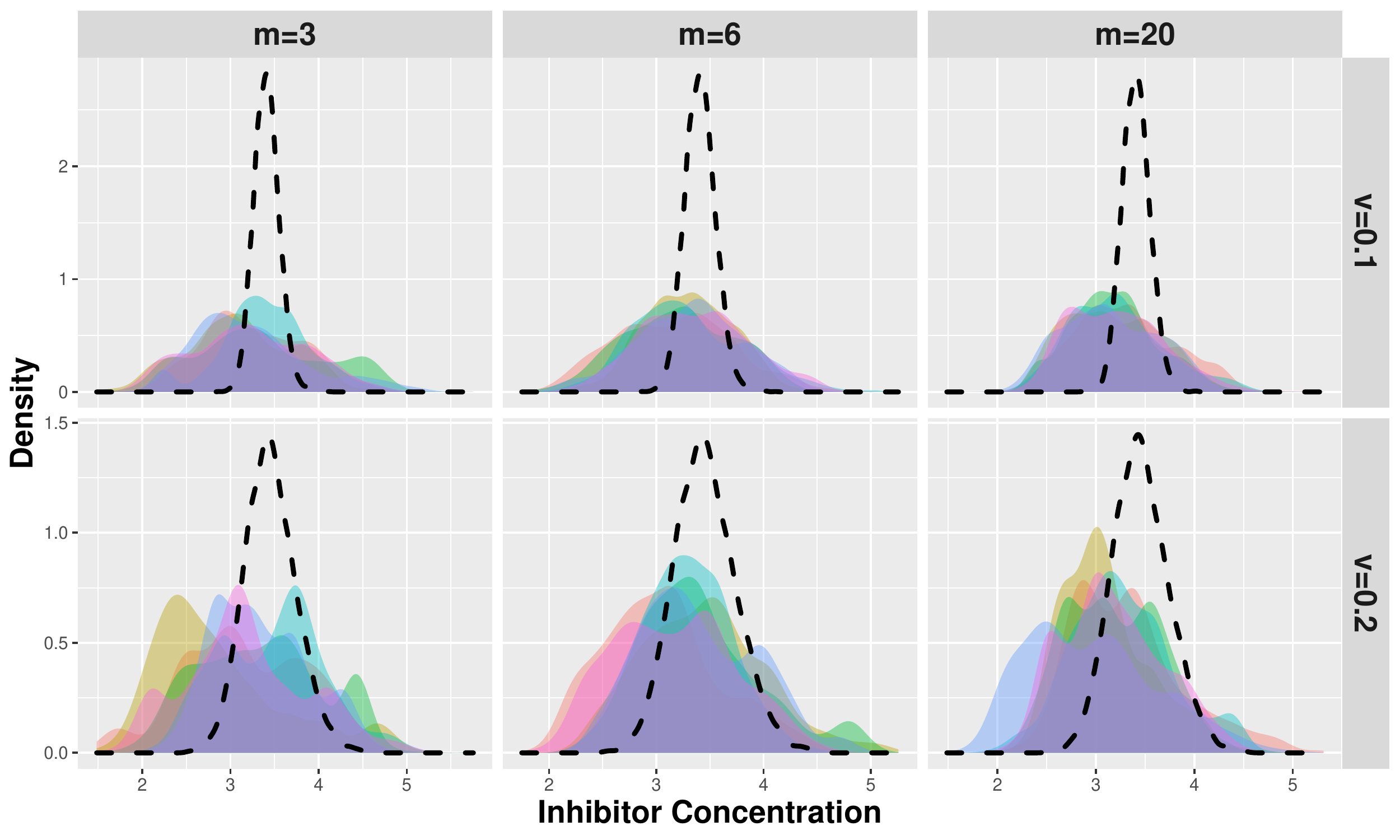}
         \label{fig: inhibitor concentration with auxilary}
     }
     \vspace{-0.2em}
     \subfloat[Cell density (without LG-DBN auxiliary)]{
         \centering
         \includegraphics[width=0.485\textwidth]{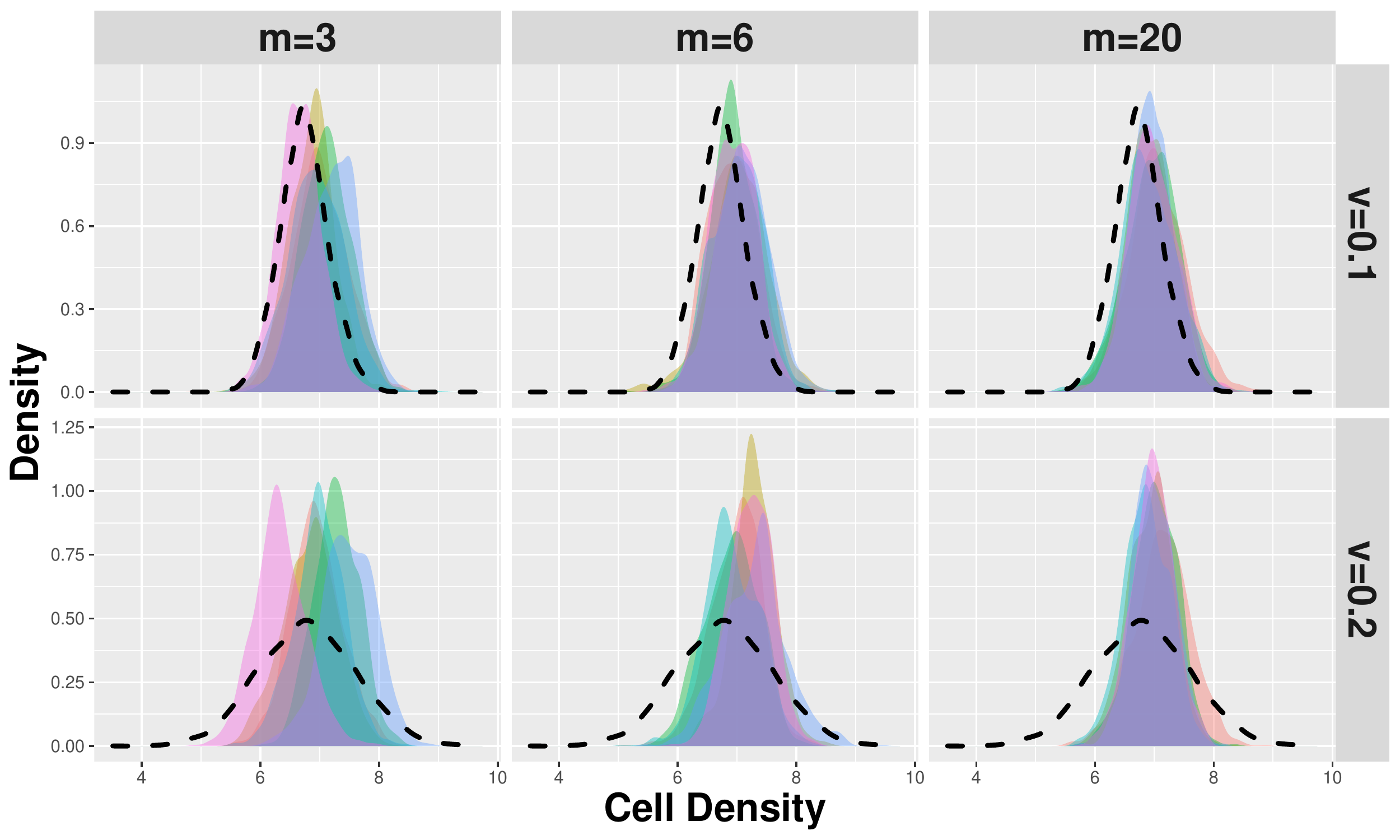}
         \label{fig: cell density without auxilary}
     }
   \subfloat[Inhibitor concentration (without LG-DBN auxiliary)]{
         \centering
         \includegraphics[width=0.485\textwidth]{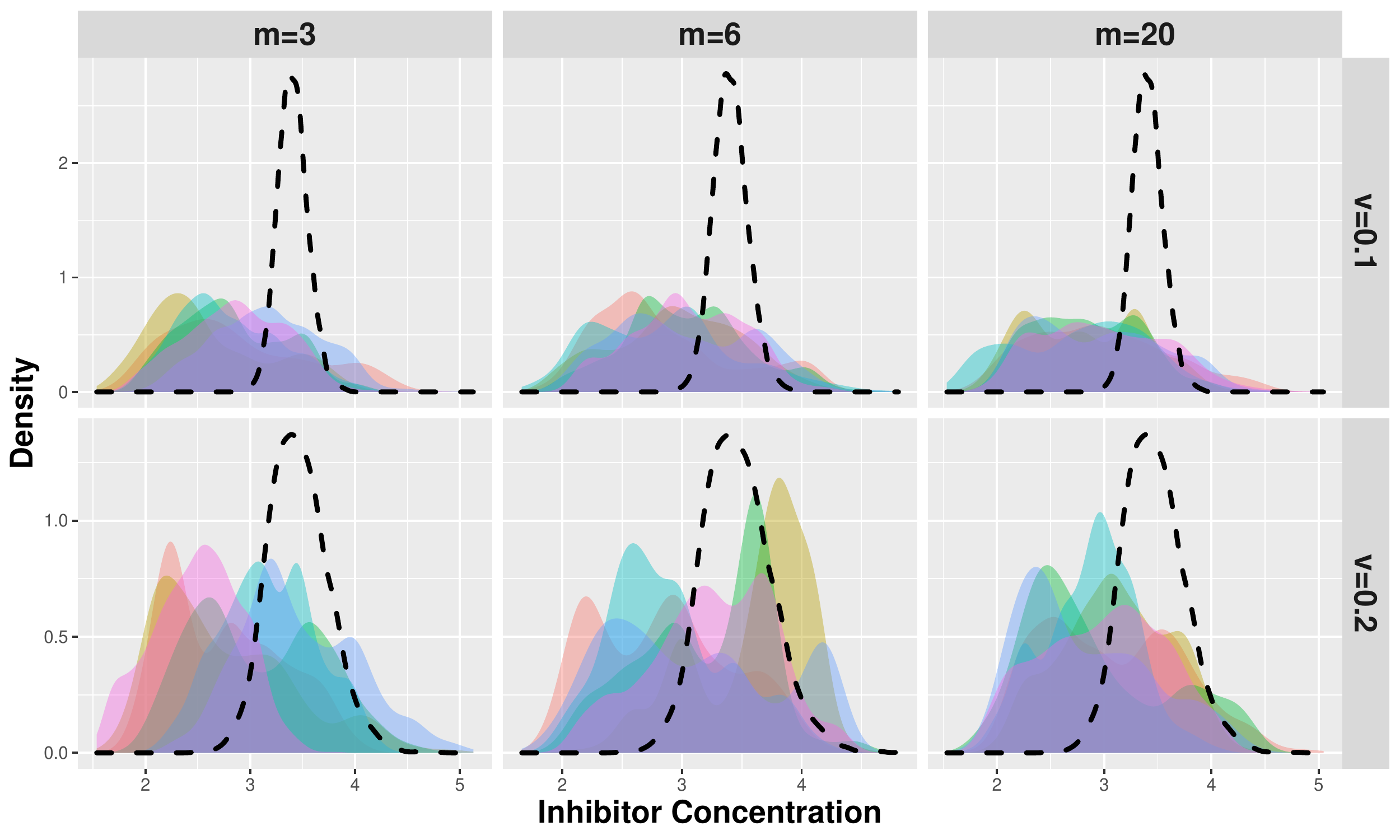}
         \label{fig: inhibitor concentration without auxilary}}
     \medskip
      \vspace{-0.1in}
     \caption{Posterior predictive distributions of cell density and inhibitor concentration at the 30-th hour ($t=11$) $p(p_{t},I_{t}|\rho_1,I_1)$ obtained from 6 macro-replications (simulated with common random numbers). The color filled areas under the probability density curve represent estimated posterior predictive distributions from different macro-replications. The black dashed line represents the predictive distribution of the ``true" model, i.e. $p(\rho_t,I_t| \rho_1,I_1, \pmb\theta^c)$. The rows of each panel are related to noise levels (i.e. $v=0.1, 0.2$) while the columns of each panel are sample sizes of observations (i.e., $m=3,6,20$). 
     }
     \label{fig: predictive distribution}
     \vspace{-0.1in}
\end{figure}

Then, we compare the prediction accuracy of the posterior predictive distribution obtained from ABC-SMC with and without LG-DBN auxiliary. We estimate the parameters $\pmb\theta=(r_g, k_s,k_c,r_d,v_\rho,v_I)$. 
Specifically, in each macro replication, we generate posterior samples $\left\{\pmb\theta^{(i)}\right\}_{i=1}^{N_\alpha}$ by LG-DBN auxiliary and naive ABC-SMC approaches to approximate the posterior predictive distribution, 
\vspace{-0.1 in}

$$p(\rho_t, I_t |\rho_1, I_1, \mathcal{D})= \int p(\rho_t, I_t|\pmb\theta, \rho_1, I_1)p(\pmb{\theta}|\mathcal{D})d \pmb\theta=\frac{1}{N_\alpha}\sum^{N_\alpha}_{i=1}p\left(\rho_t,I_t|\rho_1,I_1,\pmb\theta^{(i)}\right),$$
\vspace{-0.1in}

\noindent where the probability density $p(\rho_t, I_t|\rho_1, I_1, \pmb\theta^{(i)})$ is computed by the hybrid model \eqref{equ:hybridinhibition} for $i=1,2,\ldots,N_\alpha$. Given the ``true" model parameters $\pmb\theta^c$, we can also construct the predictive distribution $p(\rho_t,I_t| \rho_1,I_1, \pmb\theta^c)$ from the model \eqref{equ:hybridinhibition}. Figure~\ref{fig: predictive distribution} shows posterior predictive distributions of cell density and inhibitor concentration at the 30-th hour or timestep $t=11$ given a fixed initial state $(\rho_1,I_1)=(3,0)$. The black dashed line represents the predictive distribution of ``true" model $p(\rho_{11}, I_{11}|\rho_1,I_1, \pmb\theta^c)$.

\begin{table*}[th]
\caption{The K-S statistics of cell density and inhibitor accumulation at the 30-th hour (i.e., $t=11$).}
\small
\centering
\label{table: predks_rho&I}
\begin{tabular}{c|c|ccc|ccc}
\toprule
             & & \multicolumn{3}{c|}{ABC-SMC with LG-DBN auxiliary} & \multicolumn{3}{c}{ABC-SMC without LG-DBN auxiliary} \\ \midrule
State               & Process Noise & \multicolumn{1}{c|}{$m = 3$}      & \multicolumn{1}{c|}{$m = 6$}      & \multicolumn{1}{c|}{$m = 20$}      & \multicolumn{1}{c|}{$m = 3$}      & \multicolumn{1}{c|}{$m = 6$}      & \multicolumn{1}{c}{$m = 20$}         \\ \midrule
\multirow{3}{*}{$\rho_{t}$} & $v = 0.1$       &\multicolumn{1}{c|}{0.34 $\pm$ 0.04}  & \multicolumn{1}{c|}{0.31 $\pm$ 0.03}  & \multicolumn{1}{c|}{0.25 $\pm$ 0.02}   &\multicolumn{1}{c|}{0.26 $\pm$ 0.05}  &\multicolumn{1}{c|}{0.24 $\pm$ 0.04}    &\multicolumn{1}{c}{0.23 $\pm$ 0.03}   \\ \cmidrule{2-8}
            & $v = 0.2$      & \multicolumn{1}{c|}{0.25 $\pm$ 0.05}        & \multicolumn{1}{c|}{0.22 $\pm$ 0.04}     & \multicolumn{1}{c|}{0.19 $\pm$ 0.02}   & \multicolumn{1}{c|}{0.36 $\pm$ 0.04} & \multicolumn{1}{c|}{0.32 $\pm$  0.03}  & 0.28 $\pm$ 0.02   \\ \cmidrule{1-8} 
            
\multirow{3}{*}{$I_{t}$} & $v = 0.1$       & \multicolumn{1}{c|}{0.45 $\pm$ 0.04}  & \multicolumn{1}{c|}{0.46 $\pm$ 0.03} &\multicolumn{1}{c|}{0.44 $\pm$ 0.02}    & \multicolumn{1}{c|}{0.68 $\pm$ 0.04}  & \multicolumn{1}{c|}{0.69 $\pm$ 0.03}   &  \multicolumn{1}{c}{0.67 $\pm$ 0.02}   \\ \cmidrule{2-8} 
            & $v = 0.2$       & \multicolumn{1}{c|}{0.38 $\pm$ 0.05}      & \multicolumn{1}{c|}{0.37 $\pm$ 0.05}  & \multicolumn{1}{c|}{0.36 $\pm$ 0.04}    & \multicolumn{1}{c|}{0.53 $\pm$ 0.07}      & \multicolumn{1}{c|}{0.55 $\pm$  0.06}       & \multicolumn{1}{c}{0.56 $\pm$  0.04}  \\ \bottomrule
\end{tabular}
\end{table*}

By comparing Figure~\ref{fig: predictive distribution}\subref{fig: cell density with auxilary}-\subref{fig: inhibitor concentration with auxilary} to Figure~\ref{fig: predictive distribution}\subref{fig: cell density without auxilary}-\subref{fig: inhibitor concentration without auxilary}, we observe that LG-DBN auxiliary ABC-SMC shows more robust performance across macro-replications and the posterior predictive distributions are generally closer to the ``true" predictive distribution than naive ABC-SMC. 
We further investigate Panel \subref{fig: cell density with auxilary} and \subref{fig: cell density without auxilary}. In low noise level $v=0.1$, the auxiliary based ABC-SMC tends to overestimate the variance $v_\rho$ causing the estimated posterior predictive distributions more flat than the ``true" predictive distribution. However, in high noise level, the posterior predictive distribution of LG-DBN auxiliary ABC-SMC is more accurate than that from naive ABC-SMC which consistently underestimates the variance $v_\rho$. 
The LG-DBN auxiliary ABC-SMC consistently shows better prediction on inhibitor concentration; see Figure~\ref{fig: predictive distribution}\subref{fig: inhibitor concentration with auxilary} and \ref{fig: predictive distribution}\subref{fig: inhibitor concentration without auxilary}. 

We further use the Kolmogorov–Smirnov(K-S) statistics to assess the performance of LG-DBN auxiliary ABC-SMC and naive ABC-SMC. The K-S statistics quantifies the distance between 
posterior predictive distribution 
and predictive distribution of ``true" model. 
The K-S statistics is $D = \sup_s|F^c(s) - F^p(s)|$ for $s \in \{\rho,I\}$, where $F^c(s)$ and $F^p(s)$ are the empirical distribution functions of the samples from predictive distribution of ``true" model and posterior predictive distribution respectively.
The smaller value of K-S statistic means better approximation performance of posterior predictive distribution. The number of samples used to construct the empirical distribution is $K = 2000$ in each macro-replication. We summarize 95\% CIs of distances for both cell density and inhibitor accumulation at the 30-th hour, denoted by $\bar{D} \pm 1.96 \times S_D/\sqrt{30}$ in Table~\ref{table: predks_rho&I}, where $\bar{D} = \frac{1}{30} \sum_{r=1}^{30} D^{(r)}$ and $S_D = [\sum_{r=1}^{30} (D^{(r)} - \bar{D})^2/29]^{1/2}$.

\begin{figure}[!thb]
     \centering
     \subfloat[$r_g$ (with LG-DBN auxiliary)]{%
         \centering
     \includegraphics[width=0.49\textwidth]{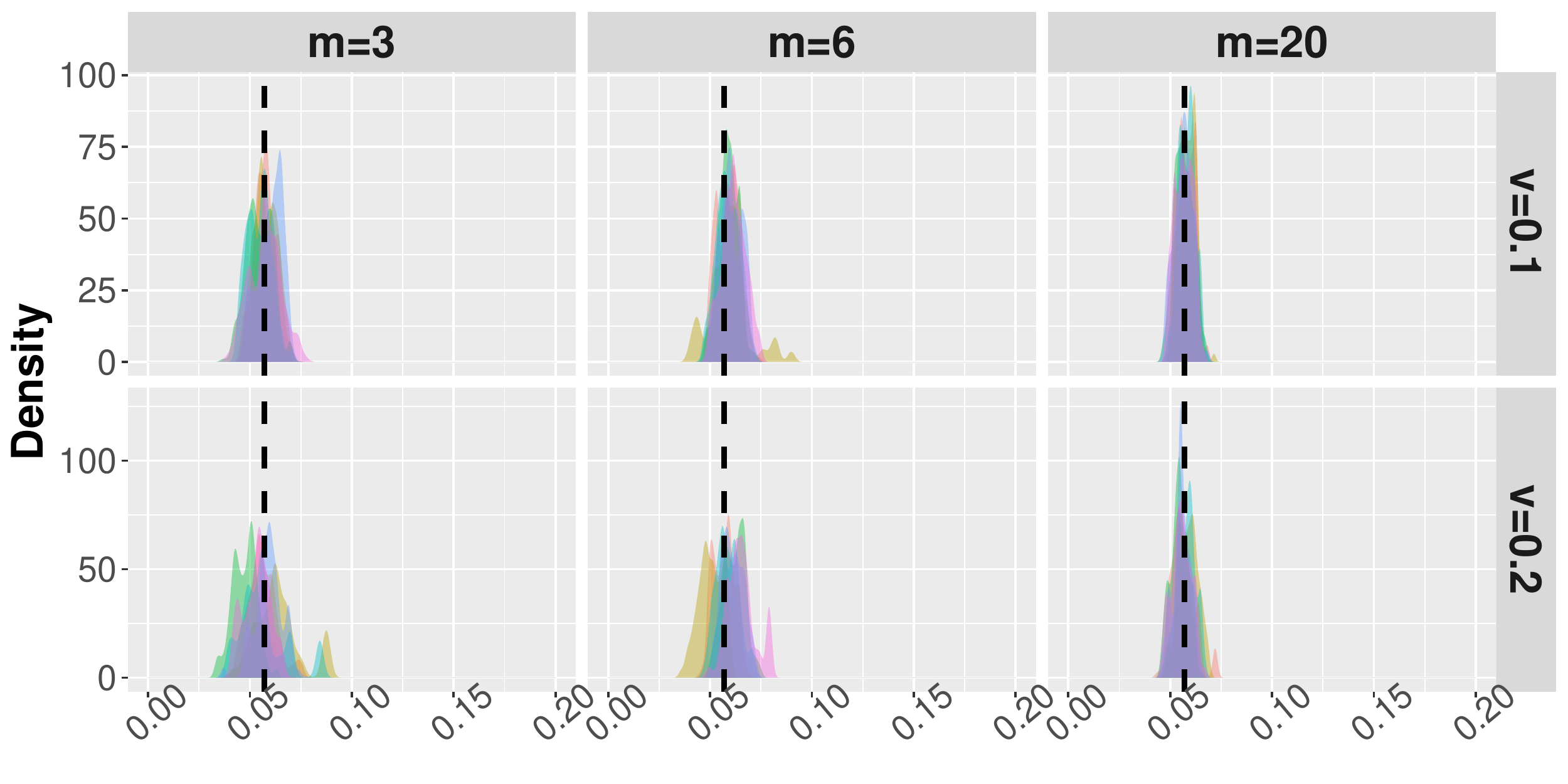}
         \label{fig: r_g_with_auxilary}}
     \subfloat[$r_g$ (without LG-DBN auxiliary)]{
         \centering
     \includegraphics[width=0.49\textwidth]{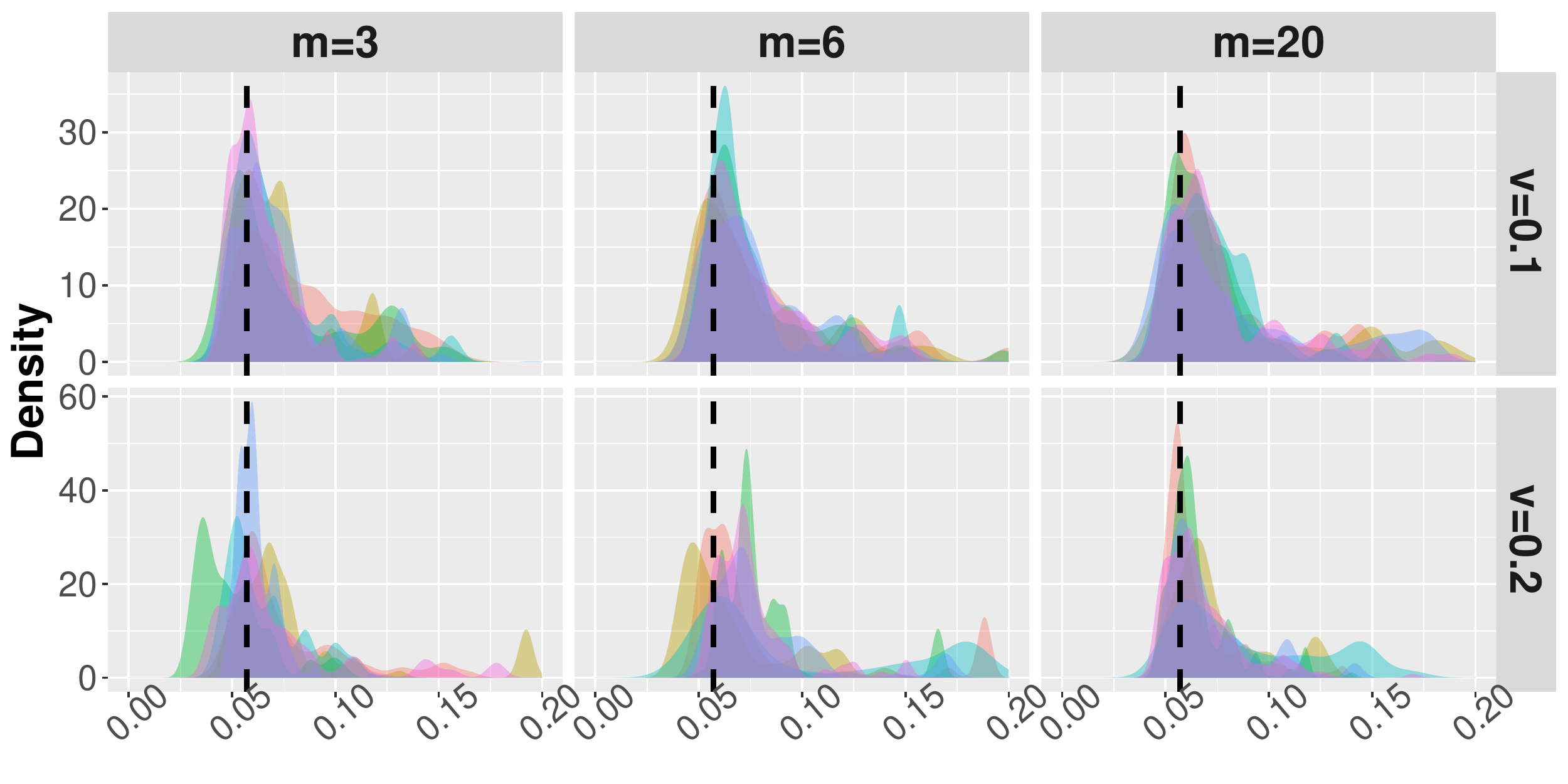}
         \label{fig: r_g_without_auxilary}
     }
    \vspace{-0.25em}
     \subfloat[$r_d$ (with LG-DBN auxiliary)]{
         \centering
         \includegraphics[width=0.485\textwidth]{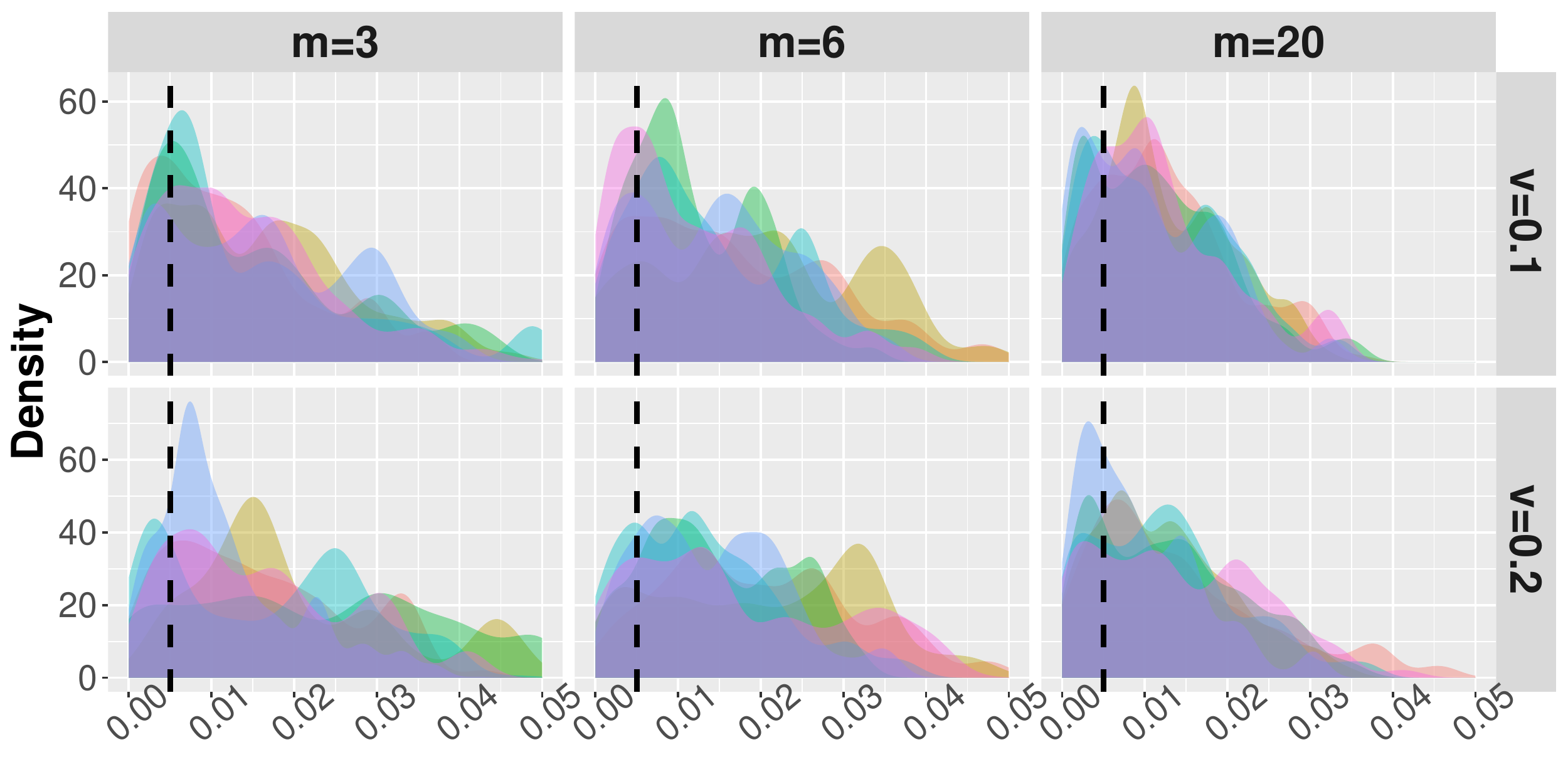}
         \label{fig: r_d_with_auxilary}
     }
   \subfloat[$r_d$ (without LG-DBN auxiliary)]{
         \centering
         \includegraphics[width=0.485\textwidth]{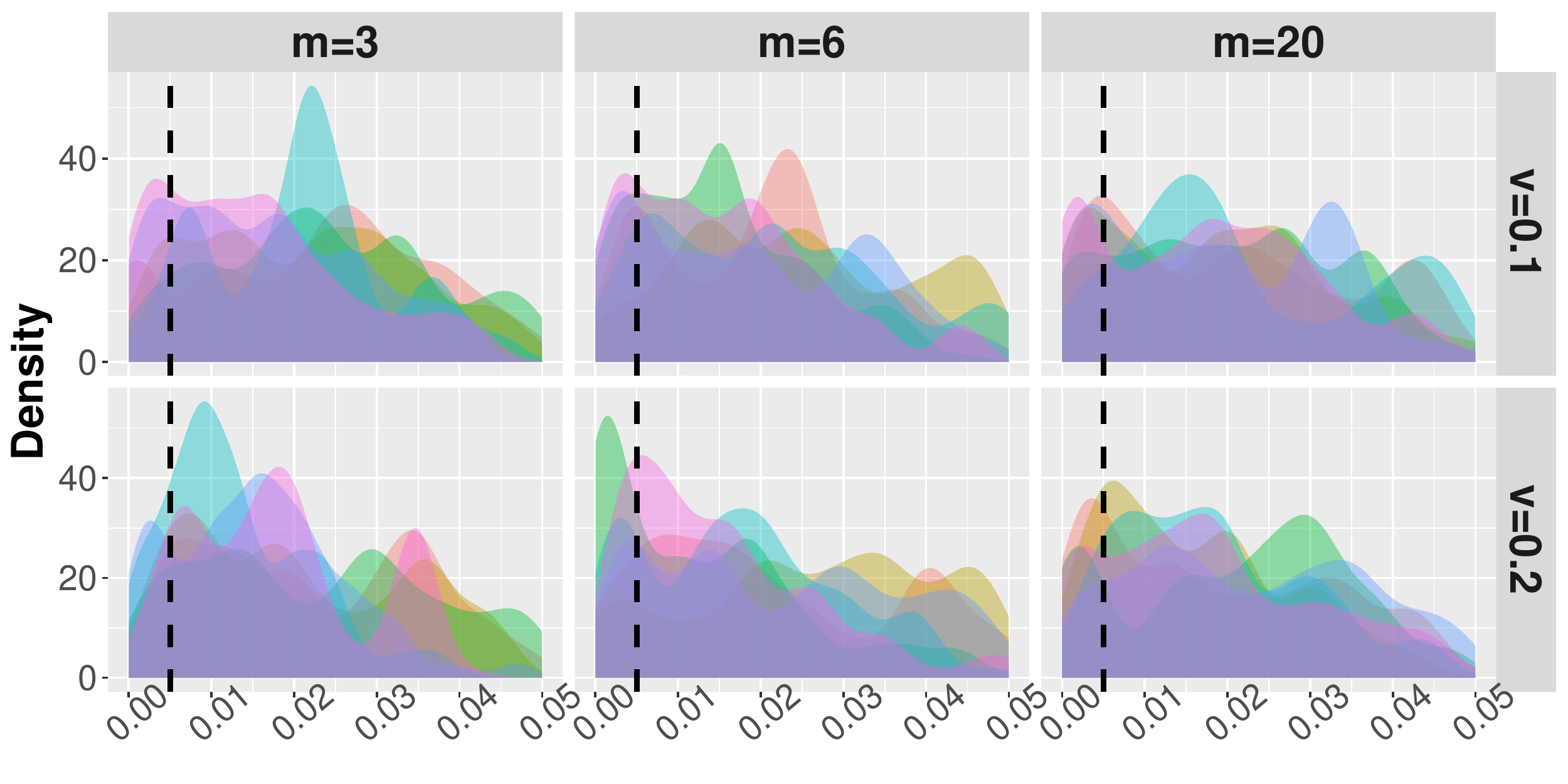}
         \label{fig: r_d_without_auxilary}}
     \medskip
      \vspace{-0.2in}
      \caption{Posterior distributions of $r_g$ and
      $r_d$ of 6 macro-replications. The posterior distributions estimated by auxiliary based ABC-SMC are shown in Panels \subref{fig: r_g_with_auxilary},
      \subref{fig: r_d_with_auxilary}. The posterior distributions estimated by naive ABC-SMC are shown in Panels \subref{fig: r_g_without_auxilary}, 
      \subref{fig: r_d_without_auxilary}.
      The black dashed lines represent the ``true" value of parameters.}
      \label{fig: posterior distribution}. 
      \vspace{- 0.2 in}
\end{figure}

\textit{As shown in Table~\ref{table: predks_rho&I}, the LG-DBN auxiliary ABC-SMC algorithm has better performance in inhibitor concentration prediction -- latent state estimation -- at all levels of model estimation uncertainty and stochastic uncertainty.} It also provides better prediction on cell density under high stochastic uncertainty. The results are consistent with the observations obtained from Figure~\ref{fig: predictive distribution}. 
The performance improvement can be further observed from the estimated posterior distribution of hybrid model parameters; see the representative plots of cell growth rate $r_g$ and inhibitor decay rate $r_d$ in Figure~\ref{fig: posterior distribution}. The posterior distribution estimated by the LG-DBN auxiliary ABC-SMC has better concentration, defined as the posterior mass around the true parameter \cite{ho2020posterior}, than naive ABC-SMC in all noise levels and sample sizes. 

Notice that due to the 
 structure of the kinetic model in (\ref{equ:hybridinhibition}) and a small value $r_d^c = 0.005$, the observable state $\rho_t$ is not so sensitive to the changes in the inhibitor decay rate $r_d$ and the inhibitor concentration $I_{t}$. Even thought it is more challenging to estimate the latent state $I_{t}$ and its mechanistic model parameter $r_d$, the LG-DBN auxiliary ABC-SMC tends to perform better.
 


\textit{In sum, compared with naive ABC-SMC, the proposed LG-DBN auxiliary ABC-SMC algorithm tends to have better prediction accuracy and computational efficiency especially under the situations with high stochastic and model uncertainties. This can benefit bioprocess mechanism learning and robust control.}
\vspace{-0.1in}

\section{\uppercase{Conclusion}}
\label{sec:conclusion}

To leverage the information from existing mechanistic models and facilitate learning from real-world data, we develop a probabilistic knowledge graph (KG) hybrid model that can faithfully capture the important properties of bioprocesses, including nonlinear reactions, partially observed state, and nonstationary dynamics.
Since the likelihood is intractable, approximate Bayesian computation (ABC) sampling strategy is used to generate samples to approximate the posterior distribution. For complex biomanufacturing processes with high stochastic and model uncertainties, it is computationally challenging to generate simulated trajectories close to real-world observations. Therefore, in this paper, we utilize a simple linear Gaussian dynamic Bayesian network (LG-DBN) auxiliary model to design 
summary statistics for ABC-SMC, which can accelerate Bayesian inference on the probabilistic KG hybrid model with high fidelity characterizing complex bioprocessing mechanisms.
The empirical study demonstrates that the proposed LG-DBN auxiliary ABC-SMC can improve computational efficiency and prediction accuracy. 
In the future research, we will extend this research to multi-scale bioprocess hybrid model in order to facilitate underlying mechanism learning, support process monitoring, and guide robust control at both cellular and system levels.
\bibliographystyle{unsrt}

\bibliography{demobib,reference,proj_ref,proposal}

\end{document}